# Method for the generation of depth images for view-based shape retrieval of 3D CAD model from partial point cloud

Hyungki Kim[1,a], Moohyun Cha[1,b], Duhwan Mun[2,c*]

[1] Division of Computer Science and Engineering, Jeonbuk National University, 567 Baekje-daero, Deokjin-gu, Jeonju-si, Jeollabuk-do 54896, Republic of Korea

[2] Dept. of Precision Mechanical Engineering, Kyungpook National University, 2559, Gyeongsang-daero, Sangju, Gyeongsangbuk-do 37224, South Korea

[a]hk.kim@jbnu.ac.kr, [b]mhcha@kimm.re.kr, [c]dhmun@knu.ac.kr

* Corresponding Author, Tel: +82-54-530-1271, Fax: +82-54-530-1278

**Abstract**

A laser scanner can easily acquire the geometric data of physical environments in the form of a point cloud. Recognizing objects from a point cloud is often required for industrial 3D reconstruction, which should include not only geometry information but also semantic information. However, recognition process is often a bottleneck in 3D reconstruction because it requires expertise on domain knowledge and intensive labor. To address this problem, various methods have been developed to recognize objects by retrieving the corresponding model in the database from an input geometry query. In recent years, the technique of converting geometric data into an image and applying view-based 3D shape retrieval has demonstrated high accuracy. Depth image which encodes depth value as intensity of pixel is frequently used for view-based 3D shape retrieval. However, geometric data collected from objects is often incomplete due to the occlusions and the limit of line of sight. Image generated by occluded point clouds lowers the performance of view-based 3D object retrieval due to loss of information. In this paper, we propose a method of viewpoint and image resolution estimation method for view-based 3D shape retrieval from point cloud query. Automatic selection of viewpoint and image resolution by calculating the data acquisition rate and density from the sampled viewpoints and image resolutions are proposed. The retrieval performance from the images generated by the proposed method is experimented and compared for various dataset. Additionally, view-based 3D shape retrieval performance with deep convolutional neural network has been experimented with the proposed method.

# 1. Introduction

Laser scanners has become a popular instrument for the 3D reconstruction of a large-scale facility, process plant, cultural heritage building or in-door environment [1] in the as-built condition. A 3D model is generated from the scanned data by collecting, pre-processing, and 3D shape modeling [2]. In the collection process, we obtain scanned data from different locations on-site to acquire point clouds. Point clouds are filtered and registered in a single coordinate system via pre-processing [3]. In the 3D shape modeling process, a 3D model with high-level semantic information such as a building information model (BIM) [4] or 3D computer-aided design (CAD) [5] is generated.

The aforementioned 3D model is generated as follows. First, points that share similar properties are separated from the point cloud. We refer to the points as a segmented point cloud [6]. Subsequently, the segmented point cloud is compared with an information object such as a door, window, or wall in the BIM or a pipe, valve, or flange in a plant 3D CAD system. The most similar information object is selected to represent the segmented point cloud and the information object itself is aligned and located in the 3D coordinate system.

As the point cloud consists of a list of 3D coordinates for points, we compare the segmented point cloud with an information object in the BIM or the catalog library of plant 3D CAD system from the viewpoint of similarity of shape and select the information object corresponding to the segmented point cloud. Thus, a 3D shape retrieval method [7, 8] is necessary such that, when the segmented point cloud is input as a query, the object that is the most similar shape is retrieved.

The characteristics of a point cloud obtained from a large-scale facility such as a building or process plant are as follows. First, it is difficult to acquire a point cloud for all the areas of the facility owing to the limitation of line of sight (LOS), even though 3D scanning is performed at multiple locations. Second, the density of points in the point cloud decreases as the distance from the 3D scanner increases.

It is very likely that the segmented point cloud of a large-scale facility does not have the overall shape of the object but only has a partial shape. Therefore, it is effective to apply a partial shape descriptor-based method [11] rather than a global shape descriptor-based method [9,10] when a segmented point cloud is input as a query and shape-based retrieval is performed to retrieve information objects from a library database (database).

Recently, among various 3D object retrieval methods based on partial shape descriptors, the view-based partial 3D shape retrieval such as [12], has achieved high retrieval accuracy. This method extracts local image features from a 2D image obtained by rendering a 3D object and compares the similarities of the extracted local features.

In view-based partial 3D shape retrieval, the viewpoint and image resolution should be known when rendering a 2D image from a segmented point cloud. Depending on the viewpoint and image resolution, the completeness of the 2D image varies. In general, the closer the rendering viewpoint is to the scanning location, loss of geometry information in rendered image decreases. However, as the information on the location of the 3D scanner is lost during processing, registering, and segmenting point clouds acquired by scanning a large-scale facility, a method for the automatic selection of viewpoint and image resolution is required.

In this paper, we propose a method for selecting the viewpoint and image resolution required to apply view-based partial 3D object retrieval to the segmented point cloud obtained from scanning a large-scale facility. In the proposed method, several depth images are rendered from the sampled viewpoints and resolutions to calculate the acquisition rate and density for each image. Subsequently, the method automatically selects the viewpoint and image resolution to achieve a high-quality depth image for view-based partial object retrieval.

The remainder of the paper is organized as follows. Section 2 reviews the related literature and studies. Section 3 describes the view-based partial shape retrieval method. In Section 4, we analyze the effectiveness of viewpoint and image resolution on the view-based partial shape retrieval and propose our method for the selection of the viewpoint and image resolution using the quantity and density measure. In Section 5, the proposed method is verified via the implementation of a prototype retrieval system and deep-learning-based retrieval experiments with CAD data sets. Section 6 concludes our study and presents future works.

## 2. Review of related literature and studies

In the comparison of shapes of 3D objects, shape descriptors representing compact features of 3D shapes are extracted from 3D objects, and the similarities between them are calculated. In [8], the shape comparison methods were classified into feature-based, graph-based, and geometry-based methods according to the approach of expressing shape descriptors.

In the feature-based method, features representing the geometrical properties of 3D shapes are calculated and compared. The representative features include global features such as volume [13] and moment [14], spatial maps such as spherical harmonics [15] or 3D Zernike [16], and local features such as point signature [17] or persistent feature histogram [18].

In the graph-based method, the graph representing the relationship among the components of 3D shapes is used to extract the geometric meanings of 3D shapes and the extracted geometric meanings are compared to calculate the similarity between 3D shapes. Model graph [19], Reeb graph [20], and skeleton [21] are representative types of graphs.

Geometry-based methods are classified as view-based method [12,22], volumetric-error-based method [23], weight-point-set-based method [24], and deformation-based method [25]. In the view-based method, a target 2D image set is generated from each model in the database in advance. The 2D image set generated from the input 3D model or a 2D sketch is used as the query model; the query model is thereafter compared to the target 2D image sets generated from the models stored in the database.

View-based methods can be applied for global and partial retrieval [8]. In global retrieval, the sum of the distances between the view pairs in the two models is used as the dissimilarity measure. In partial retrieval, the minimum of the distances between the view pairs is used as the dissimilarity measure.

In the view-based method, 3D objects in the database have a complete shape and 2D images can be directly generated by rendering 3D shapes in various predefined directions. However, when the query is a segmented point cloud extracted from the entire point cloud of a large facility, the position and direction of the viewpoint at which the segmented point cloud is rendered, along with the resolution of the 2D image, influence the quality of the 2D image. Therefore, a method for automatically selecting the optimal rendering viewpoint and image resolution for the input point cloud is required. However, there are few studies on the methods to determine the optimal rendering point and image resolution from a segmented point cloud. The scale-weighted dense bag-of-visual-features [12], which is a view-based partial retrieval method, utilizes a scale-invariant feature transform (SIFT), which is a common image feature in view-based 3D object retrieval. Our method for selecting the viewpoint and image resolution required to generate a query image is integrated into the view-based partial retrieval method. The authors of [12] used a polygon mesh as a query model. In this research, we proposed the method of viewpoint and resolution estimation method to retrieve 3D object from a point cloud query.

Recently, deep learning-based shape retrieval approaches, such as multi-view CNN [26] and RotationNet [27], have also been proposed. Input of abovementioned deep learning-based shape retrieval approaches are images rendered from complete 3D polygonal meshes, partial polygonal meshes triangulated from 3D scan or real world images. As we discussed in Section 4, triangulation process requires additional computational cost. On the other hand, other approaches such as voxel-based [28] or unordered point set-based [29] approaches are also investigated.

Viewpoint selection studies are conducted in the computer graphics research field for scene exploration, image-based rendering, and next best view application. Early researches such as [30] proposed to estimate viewpoints by viewpoint entropy, which is calculated using a proportion of face area seen from viewpoint. In [31], unified framework for viewpoint selection based on information channel is proposed. However, [30] and [31] focuses on polygonal mesh model where topology information is included in model geometry, unlike point clouds. Recent researches such as [32] address the saliency of point clouds without topology information.

However, to the best of our knowledge, adopting saliency of point clouds to viewpoint selection problem has not been studied yet. Moreover, the effectiveness of selected viewpoint on view-based shape retrieval should be further investigated.

## 3. View-based partial shape retrieval

In our research, the view-based partial shape retrieval method of [12] has been mainly selected and improved to experiment the effect of proposed viewpoint and resolution estimation selection method. In [12], depth images are rendered at various angles and a SIFT [30] feature is extracted from each image to generate a descriptor from 3D models in DB. In order to improve the accuracy of the retrieval, keypoints are extracted randomly in the foreground region where an object exists, in contrast to the conventional keypoint detection of SIFT. Moreover, in order to perform robust retrieval from a query model that may have discontinuous geometry owing to incomplete scanning or noise, keypoints are detected from a multiresolution image pyramid with a similar density to obtain a low-pass filtering effect. The experiments in [12] demonstrate that extracting 3,600 keypoints based on 256 × 256 depth images yields the best retrieval performance.

View based partial shape retrieval approach compares 3D objects using rendered images. To compare the two images, image descriptor is encoded from obtained features. In [12], authors adopted hard assignment encoding of bag-of-visual-words method. However, the expression power of descriptor can be increased by using soft assignment, vector of locally aggregated descriptors (VLAD) [34] or Fisher vector [35]. We adopted Fisher vector method for descriptor encoding, which encodes the difference between the Gaussian Mixture Model (GMM) constructed from the feature set of database and a set of features from an image. Constructed GMM is represented by $K$ number of weight, mean and variance parameters $\lambda =$

$\{w_k, \mu_k, \sigma_k, k = 1, \ldots, K\}$. For each image has n number of features where each feature is D-dimensional, gradient with respect to μ and σ can be expressed as Equation (1) and (2)

$$u_k = \frac{1}{n\sqrt{w_k}} \sum_{i=1}^{n} q_k(x_i) \frac{x_i - \mu_k}{\sigma_k} \tag{1}$$

$$v_k = \frac{1}{n\sqrt{w_k}} \sum_{i=1}^{n} q_k(x_i) \frac{1}{\sqrt{2}} \left[ \frac{(x_i - \mu_k)^2}{\sigma_k^2} - 1 \right], \tag{2}$$

where $x_i$ is a D-dimensional features of image and $q_k(x_i)$ is the soft assignment of $x_i$ to Gaussian k. The final descriptor of an image is $d = [u_1, v_1, \ldots, u_K, v_K]$, which is 2DK-dimensional vector. The retrieval is conducted by calculating the distance between descriptor of the query image with the descriptors of the database images where smaller distance indicates similar image.

Figure 1 shows a process designed based on the abovementioned approach, where we considered a segmented point cloud as the input query and output is descriptors using Fisher vector encoding. The segmented point clouds in database are pre-processed in offline to generate descriptor database. Note that in case of large facilities such as process plants, most of the objects are defined in a catalog. The standardized catalog is a data model that stores CAD models for common reusable parts. Thus, the execution of the offline process does not require frequent updates. In reconstruction process when point cloud is captured and segmented, user needs to find corresponding model type in database. The retrieval is an online process that generates a descriptor from a segmented point cloud input as a query, determines the most similar model by comparing the query descriptor with the descriptor database, and returns the model as a retrieval result. Note that Gaussian Mixture model estimation is performed only in offline process and estimated GMM parameters are reused in online retrieval process.

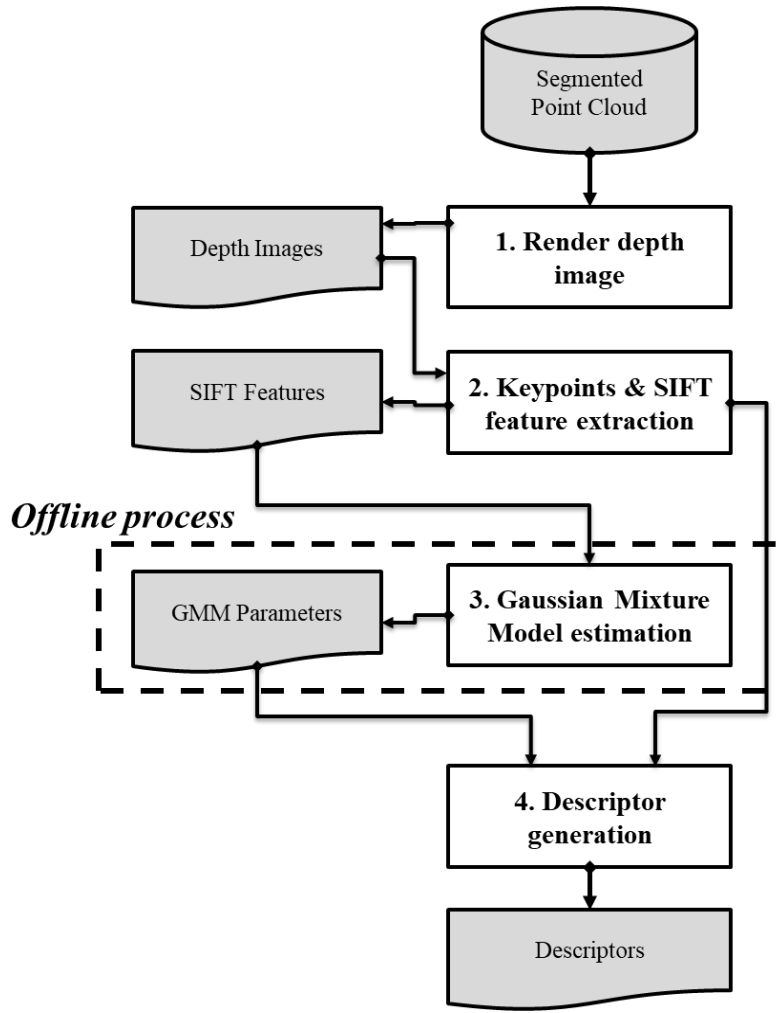

**Figure 1.** Designed process for view-based partial shape retrieval

In the depth image rendering process, position and scale of the input segmented point cloud is normalized and depth images are rendered. In the position and scale normalization, the center of mass of segmented point cloud is calculated and the translation is performed to all point data such that the center of mass of segmented point became the global origin. Subsequently, scaling was performed such that the distance between the coordinates of the farthest point and global origin was one. Therefore, a unit sphere centered at the global origin enclosed the input model.

After the normalization of position and scale, viewpoint and resolution should be decided before depth image rendering. As shown in Figure 2, depth image is a rendered image where the intensity of pixel indicates the distance of geometry from viewpoint. The closer the distance, the larger the value (brighter color), and vice versa. The detailed process for deciding viewpoint and image resolution is explained in next section.

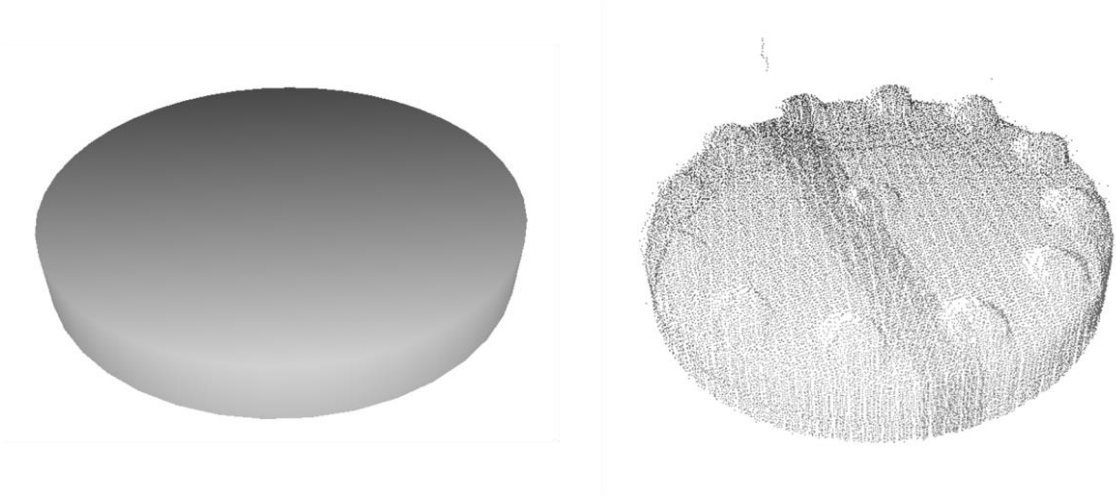

**Figure 2.** Rendered depth image representing brighter color if the geometry is closer to the viewpoint (left) example depth image for polygonal mesh model (right) depth image rendered for segmented point cloud.

In the keypoints and SIFT feature extraction process, the keypoints from the rendered depth images are detected and the SIFT image features at each keypoint is calculated. The SIFT algorithm detects the salient pixels as keypoints and generates 128 dimension features from the multiresolution image pyramid using the gradient histogram near each keypoint. As described in [12], the number of pixels (keypoints) detected by the keypoint detection method proposed in [33] was insufficient for retrieval. Therefore, we added a module to select pixels randomly from the foreground region of the image where the shape exists in the depth image, as in the method proposed in [12], and calculated the features of the selected keypoints.

The parameters used in the keypoints and SIFT feature extraction process are $N_k$ and S. $N_k$

is the number of pixels randomly sampled from the rendered depth image. $S$ is the reduction ratio of the pixels to be sampled as the size of the images in the multiresolution image pyramid decreases. In the multiresolution image pyramid, when the size of the image is reduced to half in width and height, the number of pixels to be sampled is reduced to $N_k/S$. In our retrieval system, $N_k$ was selected as 1200 and $S$ was selected as 1.5, as these values demonstrated the best results in the experiment performed in [12].

In the Gaussian mixture model estimation process, entire SIFT features detected from the rendered depth images of point clouds in the database is considered. As described before, descriptor encoding using Fisher vector results 2DK-dimensional descriptor where D indicates feature dimension, which is fixed as 128 since we adopted SIFT image feature. Therefore $K$ is a user-provided parameter for Gaussian mixture model estimation. To refer parameters obtained from Gaussian mixture model estimation process in offline, $\lambda = \{w_k, \mu_k, \sigma_k, k = 1, ..., K\}$ values are stored as GMM parameters. GMM parameters are reused in online process to encode descriptor from an image. The descriptor generation process is a process of encoding SIFT features obtained from the depth image into a single descriptor using the GMM parameter. Using Equation (1) and (2), one 2DK-dimensional descriptor is obtained for each depth image.

In online process where segmented point cloud is given to retrieve corresponding type, the same process is performed to query. For query point cloud, descriptors are generated for each rendered depth image. Then the distance between descriptors from query point cloud and each descriptor in descriptor database is calculated. The distance metric is the standard cosine distance between two descriptors.

# 4. Generation of depth image from segmented point cloud

## 4.1 Problem of selection of viewpoint and image resolution for rendering

As mentioned above, the purpose of this study is to retrieve a type existing in the database by segmented point cloud query obtained from scanning large-scale facilities as the query. The query model assumed in our research has the following problems related to the view-based partial shape retrieval.

1) *Using point cloud as a query*: The scan data are discrete data having a smaller amount of information than the original shape because a point cloud is the sampling result from the surface. Converting a discrete point cloud into a continuous surface is possible by performing surface reconstruction [36] as a post-process. However, the computational cost of surface reconstruction is high and the error owing to the reconstruction is incurred in addition to the error owing to the scanner noise.

2) *Point cloud without information on scanner locations*: Scanners that encode the global location of a device are very limited. Therefore, a retrieval system must provide viewpoint information manually, which is a limitation on the usability and scalability of the retrieval system. If the scan data contain the result of a single scan frame, it can be assumed that the origin of the scan data is the position of the scanner. However, as our query is a segmented point cloud separated from the entire point cloud, which is the registered point cloud from multiple scan locations, there is no information about the locations of scanners.

3) *Point cloud without normal*: Only structural light or stereo vision-based scanning has a possibility to provide additional information for normal estimation. In the case of a long-range scanner such as LiDAR, the normal of each point should be estimated via additional post-processing. Furthermore, most point normal estimation methods require the approximate location of the scanner.

The above problems are elaborated in the remainder of this section. Suppose that a part is installed in a facility and the scan data are collected from multiple locations. Subsequently, the registration is performed to obtain the entire point cloud for the facility. If we subsequently perform segmentation on the entire point cloud for the region of interest, depending on the location of the scanner, the geometrical information is missing due to the occlusion such as in Figure 3.

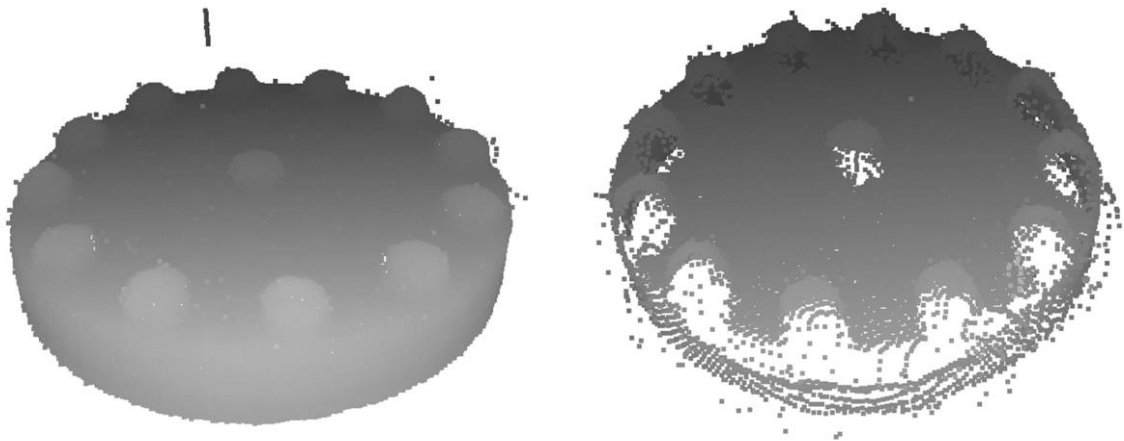

**Figure 3.** Example of a segmented point cloud; (left) image rendered near scanner direction; (right) image rendered at another viewpoint

If the segmented point cloud data are used as a query for view-based partial shape retrieval, they must be converted into an image using a depth image rendering module. The two parameters to be determined during the rendering process are the viewpoint and resolution of the image.

In order to convert 3D point cloud data into 2D image data, 3D geometry should be transformed using model, view, and projection matrixes. A model transformation transforms a 3D geometry from a local coordinate system into a global coordinate system, and a view transformation transforms a 3D geometry from a global coordinate system to a camera coordinate system. Finally, using the projection matrix, the 3D geometries of the camera coordinate system are projected onto a 2D plane. For the depth image rendering module designed in this study, the position and scale were normalized, and orthogonal projection was used such that all the data in the unit sphere belonged to the field of view. Therefore, the model and projection transformation matrix were fixed. However, the specification of the viewpoint of the camera in view transformation was challenging. If we perform position normalization on the segmented point cloud, all the points will be located around the global origin. If the viewpoint is determined, we can define the camera orientation to focus on the global origin. Subsequently, the segmented point cloud was converted into normalized device coordinates using the model, view, and projection matrixes. After the normalized device coordinates were determined, the segmented point cloud data were finally converted to image coordinate according to the designated image resolution. Therefore, the image resolution also needs to be determined.

Figure 4 illustrates the effect of the selected viewpoint and resolution to convert the segmented point cloud into a depth image. In Figure 4, for a given segmented point cloud two different viewpoint and three image resolution was selected to render depth image from point cloud. The viewpoint where camera is located, determines *quantity* of point cloud projected to image. Depending on the viewpoint, amount of self-occlusion, where some point is not projected onto the depth image because of other points in front of them, differs. The image resolution determines how *dense* the projected point appear on image as pixels through rasterization process. If targeted image resolution is low, result image has higher change to form dense and continuous image of point cloud. On the other hand, if targeted image resolution is high, projected point cloud in image forms discontinuous pixels. Continuous image is preferred since SIFT feature encodes feature based on gradient, which is not accurately calculated with discontinuous pixels.

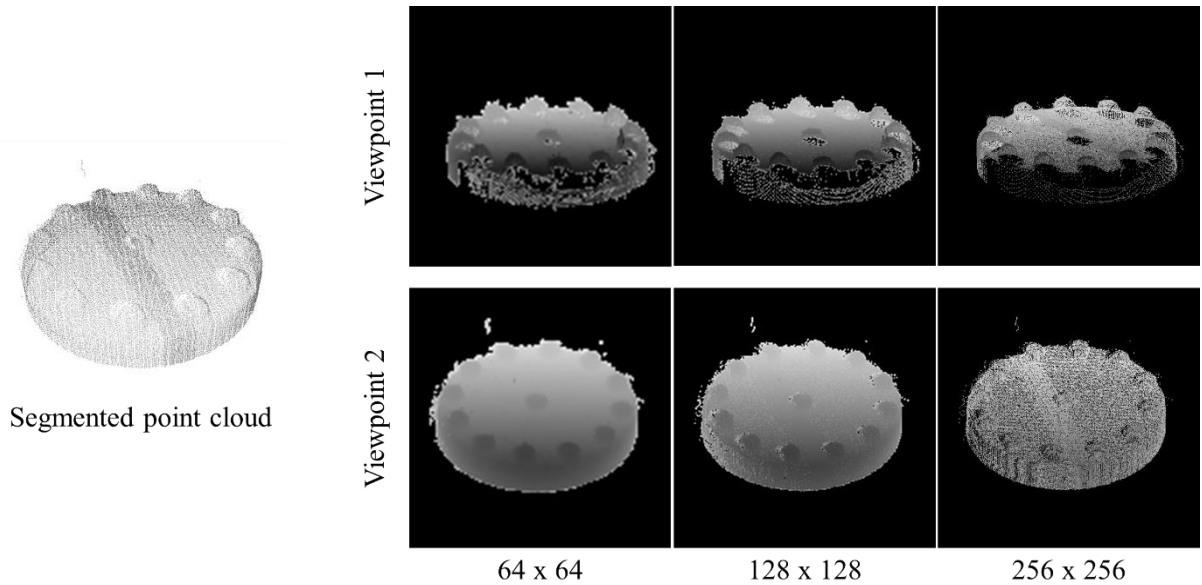

**Figure 4.** Differences between rendered depth images according to the viewpoint

It is possible to maximize the information quantity by estimating the viewpoint close to the position of the sensor from a segmented point cloud and render depth image at the viewpoint. However, the query desired in this study lost information on sensor locations during the registration and segmentation processes. Moreover, appropriate resolution of the depth image can be estimated if we have information on the scanner locations, scanner resolution, and the distance between an object and the scanner. However, this information has also been lost. Therefore, it is necessary to determine the viewpoint and appropriate image resolution for a segmented point cloud for view-based partial shape retrieval.

### 4.2 Proposed method

In the proposed method for the selection of viewpoint and image resolution, multiple depth images are rendered from sampled viewpoints in sampled resolutions, and subsequently, the

*quantity* Q and *density* D are calculated for each image. The best viewpoint and image resolution are determined based on the information quantity and density of depth image. The proposed method is designed to estimate the viewpoint such that maximum possible point of the segmented point cloud is projected into the pixels of the image. The image resolution is also determined such that pixels of projected point cloud is densely positioned.

For a segmented point cloud *P*, the *quantity* Q and *density* D are functions of viewpoint *v* and resolution *r*:

$$Q(v,r) = \frac{N_{pixel}}{|P|}, \text{ where } N_{pixel} = |I(v,r) > 0| \text{ and} \tag{3}$$

$$I(v,r) = MVP(v,r) * P,$$

$$D(v,r) = \frac{N_{8-connected}}{N_{pixel}}, \text{ where } N_{8-connected} = \left| I_B(v,r) \otimes \begin{bmatrix} 1 & 1 & 1 \\ 1 & 1 & 1 \\ 1 & 1 & 1 \end{bmatrix} = 9 \right|, \tag{4}$$

where $P$ is a segmented point cloud and $MVP(v,r)$ is a model-view-projection matrix at viewpoint *v* with targeted image resolution *r*. $v \in V = \{v_1, ..., v_{20}\}$ are viewpoints which is vertex positions of unit dodecahedron centered at origin, as illustrated in Figure 5. $r \in R = \{r_1, ..., r_{N_r}\}$ are resolutions for a depth image given by the user. $N_{pixel}$ is the number of foreground pixels of the rendered depth image *I* induced by model, view and projection matrix *MVP* defined by viewpoint *v* with resolution $r \times r$; pixels have value greater than 0 are called foreground pixels. $|P|$ is the cardinality of point cloud. $N_{8-connected}$ is the number of 8-connected pixels in the binary image $I_B$ which is binary image converted from *I*. We can obtain $N_{8-connected}$ by counting pixels that has value 9 after convolve $I_B$ with 3 by 3 matrix with 1s. Convolution operation is widely used in the image processing [35,36,37,38] research field to obtain characteristics of pixel. In Equation (4), we applied convolution to check whether a point cloud is continuous under given projection. In summary, Equation (3) represents the number of pixels onto which a segmented point cloud is projected. Equation (4) is the ratio of dense points to pixels (8-connected pixels) projected onto the depth image.

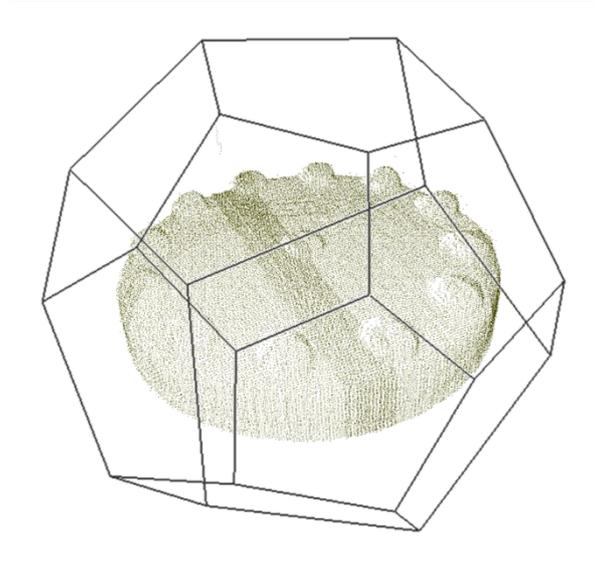

**Figure 5.** Unit dodecahedron centered at origin with normalized segmented point cloud

Figure 6 illustrates the general tendency of quantity $Q$ according to the viewpoints and resolutions. When lower the resolution, the more the quantity tends to decrease. This is because, as shown in Figure 7, the lower the resolution of the image, the more likely nearby points are projected onto the same pixel position, which results in aliasing. As each pixel can encode a single depth value, when multiple points in a point cloud correspond to the same pixel position, only the depth value of the nearest (minimum depth) point is reflected as the pixel intensity. If the resolution of the depth image is high, there is a higher possibility that depth values for nearby points are recorded into the pixels. However, at low resolution, only the depth value of a specific point among the nearby points is reflected, and the depth values for the remaining points are ignored. Therefore, the number of points to be projected is generally reduced at low resolution.

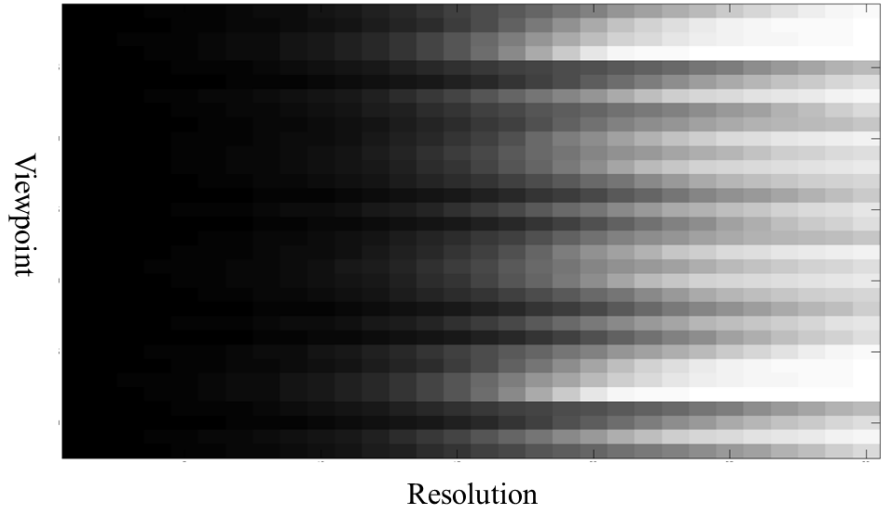

**Figure 6.** Acquisition rate according to viewpoint and image resolution

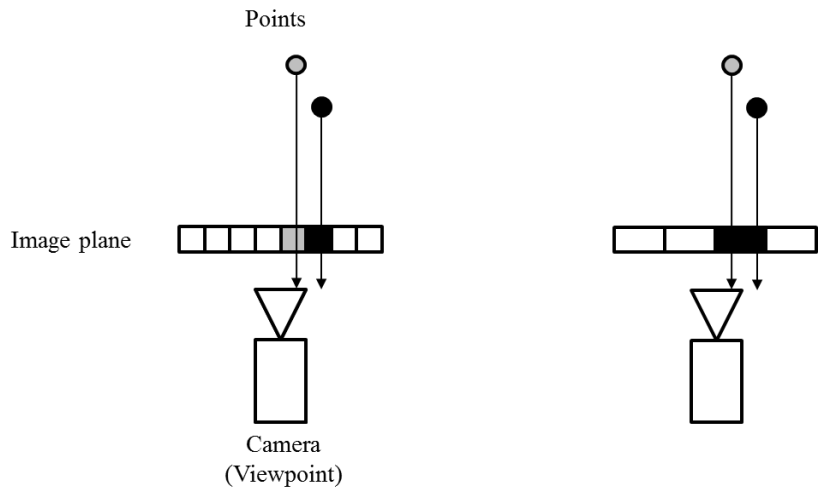

**Figure 7.** Difference in the number of projected points according to the target image resolution; (Left) high resolution; (Right) low resolution

The viewpoint is determined based on the normalized quantity, which is formulated as Equation (5). The quantities are normalized to [0, 1] across the images from the same resolution group. The result viewpoint $v^{Best} \in V$ is determined based on the sum of normalized acquisition rates for each viewpoint; we select the viewpoint $v^{Best}$ with the maximum value formulated as Equation (6). Figure 8 shows a graphical representation of the normalized quality, with the graph on the right representing the sum of normalized quality for each viewpoint.

$$\text{Normalized\_Q}(v,r) = \frac{A(v,r) - \min_{x \in W}(A(v,x))}{\max_{x \in W}(A(v,x)) - \min_{x \in W}(A(v,x))} \tag{5}$$

$$v^{Best} = \underset{v \in V}{\text{argmax}} \sum_{r=r_1}^{r_{N_r}} \text{Normalized\_Q}(v,r) \tag{6}$$

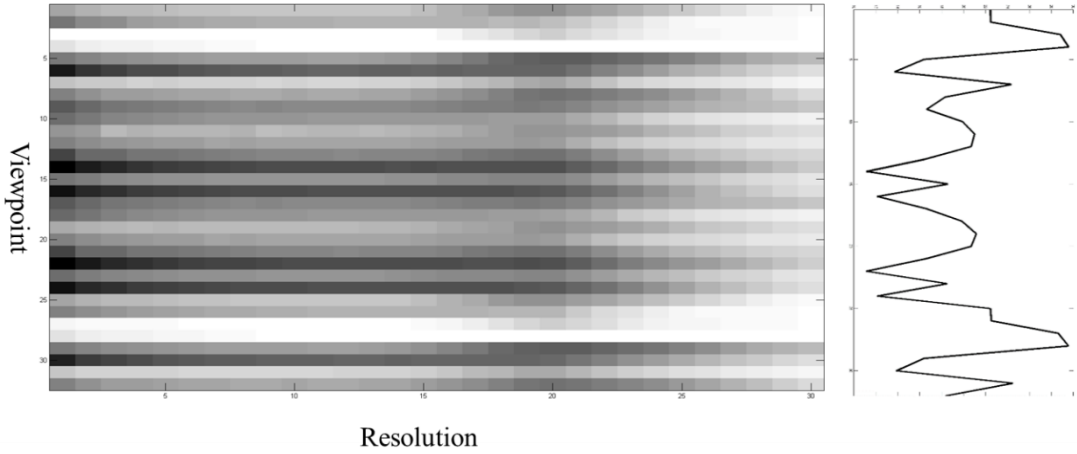

**Figure 8.** Normalized quality and sum of the rates across each viewpoint

At last, the resolution $r^{Best} \in R$ at which the density becomes maximum an the best viewpoint $v^{Best}$ is determined as the best resolution. This can be expressed by the following equation.

$$w^{Best} = \underset{r \in R}{\text{argmax}}\, D(v^{Best}, r) \tag{7}$$

For a segmented point cloud query in the on-line process, if the viewpoint and image resolution are selected using the proposed method, the view-based partial shape retrieval method can be performed with minimum loss of point cloud data without prior information about the scanning location, resolution, and distance between the target object and scanning location.

## 5. Implementation and experiment

### 5.1 Experiment on view-based partial retrieval using SIFT feature and Fisher vector

We implemented the retrieval system and conducted an experiment to observe the performance of the view-based partial shape retrieval method and the proposed method using a segmented point cloud as a query. Each of the modules described in Section 3 for view-based partial shape retrieval was implemented based on C ++ and OpenGL graphics API was used for rendering.

The first two test dataset used in our experiments were a polygonal mesh test dataset composed of simplification dataset and fitting part dataset shown in Figures 9, 10. All the models in the polygonal mesh test dataset were CAD models represented as triangular mesh. In polygonal mesh test cases, the model simplification dataset consists of 16 CAD models such that two different models exist for each of the eight part types. The two different models are the high level-of-detail (LOD) model and the low LOD model, which are similar in overall shape, and there different in small shapes such as the shapes generated by hole, pocket, and fillet. The fitting part dataset is a dataset of eight fitting part types; there are two to nine models with slightly different dimensions for each part. The scan data query for the experiment were a synthesized point cloud created from the above CAD models. The surfaces within the LOS were sampled as the points from CAD models at arbitrary scan positions. By adjusting the resolution of the sampling, we could simulate the amount of information collected for each shape.

The second test case used segmented point cloud dataset shown in Figure 11. Several point cloud is obtained by scanning plant facility. Set of points belonging to a single part is manually segmented and corresponding type in database is classified. The segmented point cloud dataset is consisting of 72 segmented point clouds where 6 segmented point cloud exist for each 12 type.

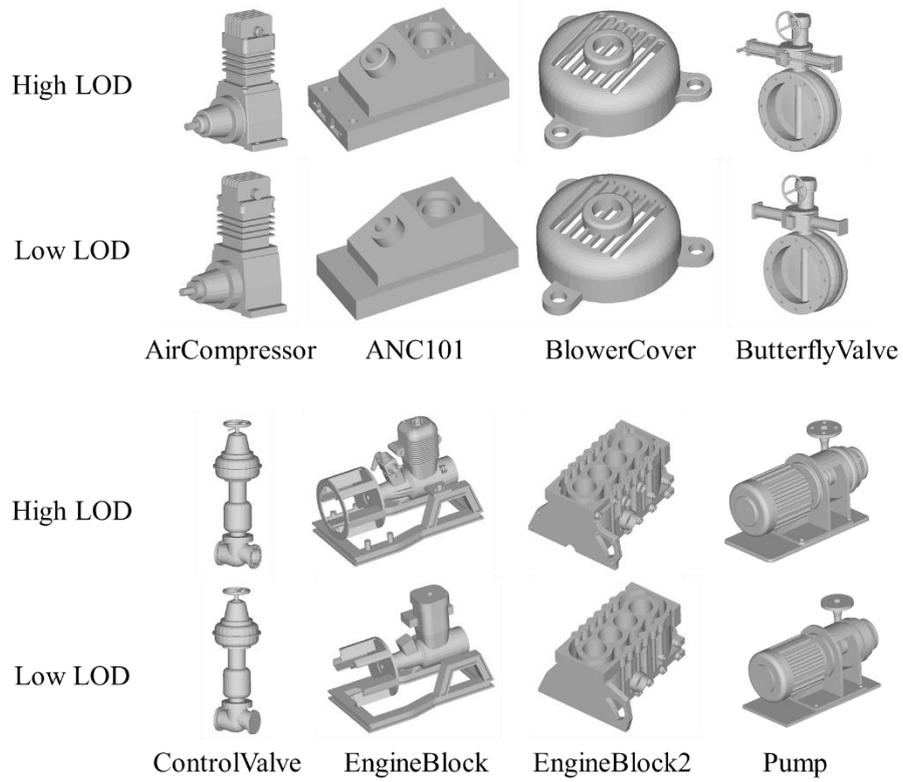

**Figure 9.** Simplification dataset of polygonal mesh test case

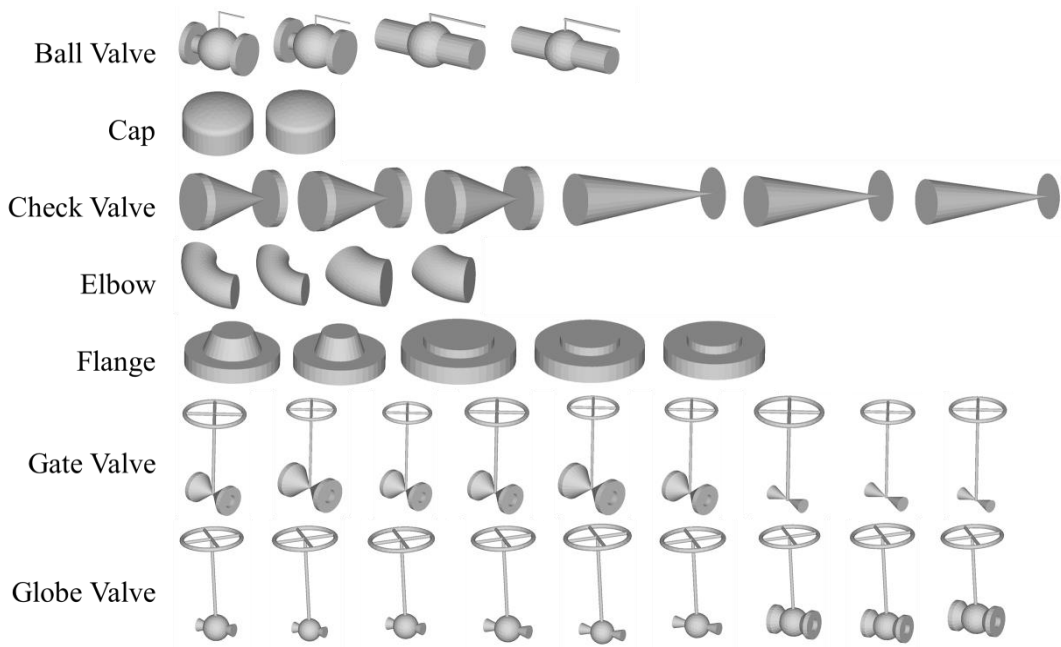

**Figure 10.** Fitting part dataset of polygonal mesh test case

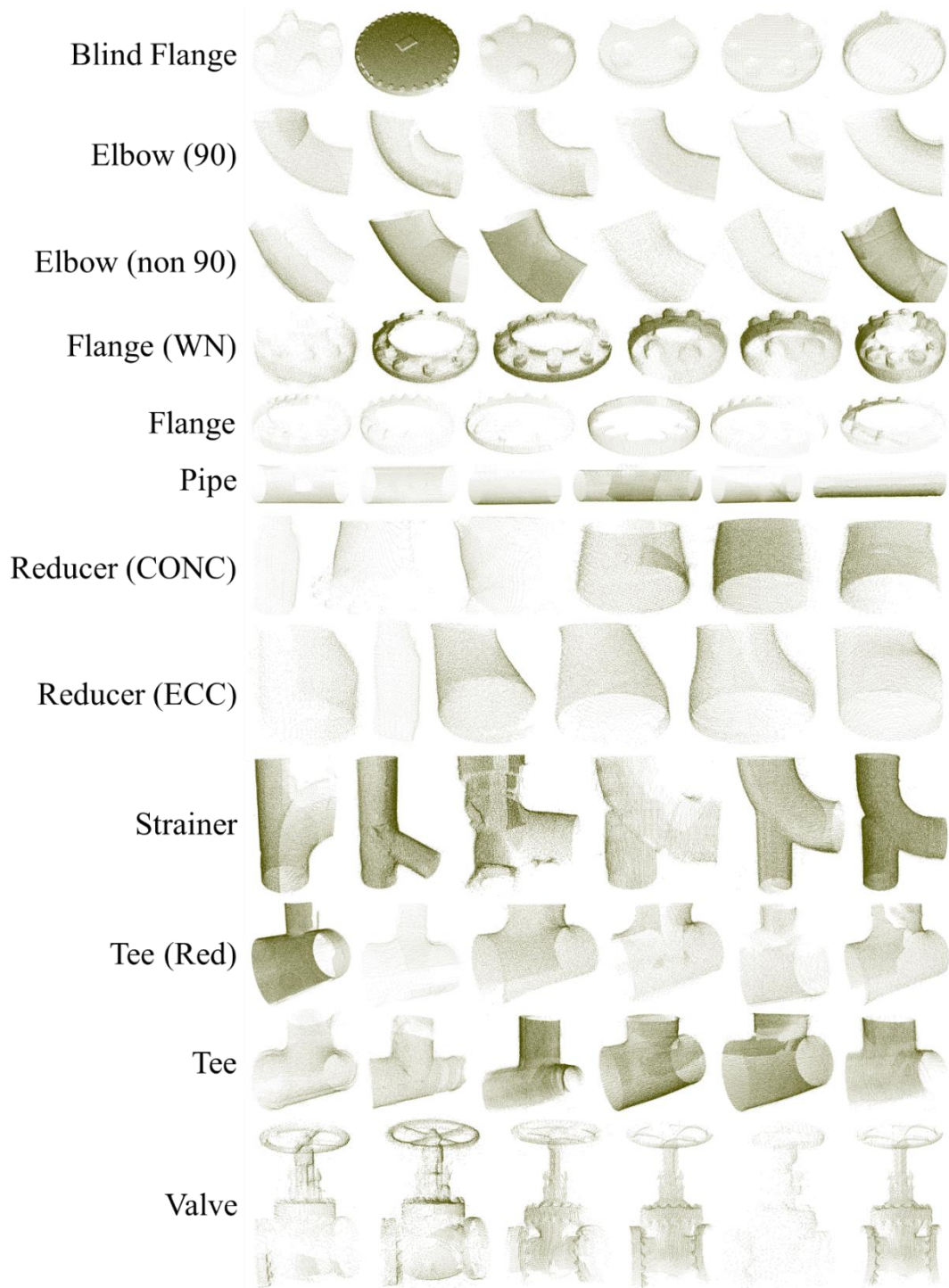

**Figure 11.** Segmented point cloud dataset

The parameters used for experiment are listed in Table 1. For the detailed meaning of each parameter, refer to Section 3. The parameters related to the proposed method are listed in Table

2. The same parameters as those in Table 1 are applied in the process of generating descriptors after selecting the viewpoint and image resolution. For the detailed meaning of each parameter, refer to Section 4.2. Selected parameter values are found before the experiment, by conducting parameter study on overall test dataset.

**Table 1** List of parameters for experiment

| Parameter | Value | Description |
|---|---|---|
| $N_k$ | 1000 | Number of sampled keypoints on the first image pyramid |
| S | 1 | Keypoints reduction ratio |
| K | 256 | Number of Gaussians |

**Table 2** List of parameters for selecting viewpoint and image resolution

| Parameters | Value | Description |
|---|---|---|
| $N_v$ | 20 | Number of view point in search space |
| $N_r$ | 8 | Number of resolution in search space |

In order to compare the retrieval accuracy according to the viewpoint and image resolution for depth image rendering, six test cases were prepared as Table 3. There are three approaches to select the viewpoint. First, the ground truth viewpoint used when generating the synthesized point cloud can be directly applied. Note that for segmented point cloud dataset, ground truth viewpoint does not exist. Therefore, test case 1 and 2 cannot be applied to segmented point cloud dataset. Second, the point cloud is fitted to the plane by random sample consensus

(RANSAC) algorithm and a viewpoint can be subsequently decided by plane normal. After position and scale normalization of segmented point cloud, three points are randomly selected to define fitting plane. Then we count the number of inlier points in segmented point cloud which is inside inlier tolerance. Iterate the process (1,000 iterations in our experiment) to find the fitted plane which has most inlier points. Then the normal of a plane induces estimated viewpoint as illustrated in Figure 12. Third, a viewpoint can be selected using the proposed method. There are two ways of choosing the image resolution. The first is to apply a fixed resolution (256 × 256 in the experiment). The second is to select a resolution using the proposed method. The resolution of 256 × 256 was found before experiment by conducting parameter study.

**Table 3.** Test cases for viewpoint and image resolution for depth image rendering

| *Test Cases* | *Test Dataset* | *Viewpoint* | *Resolution* |
|---|---|---|---|
| Case 1 | Simplification/ fitting part dataset | Ground truth | Proposed method |
| Case 2 | Simplification/ fitting part dataset | Ground truth | Fixed |
| Case 3 | All dataset[1)] | Proposed method | Proposed method |
| Case 4 | All dataset | Proposed method | Fixed |
| Case 5 | All dataset | RANSAC | Proposed method |
| Case 6 | All dataset | RANSAC | Fixed |

[1)]All dataset is simplification, fitting part and segmented point cloud dataset

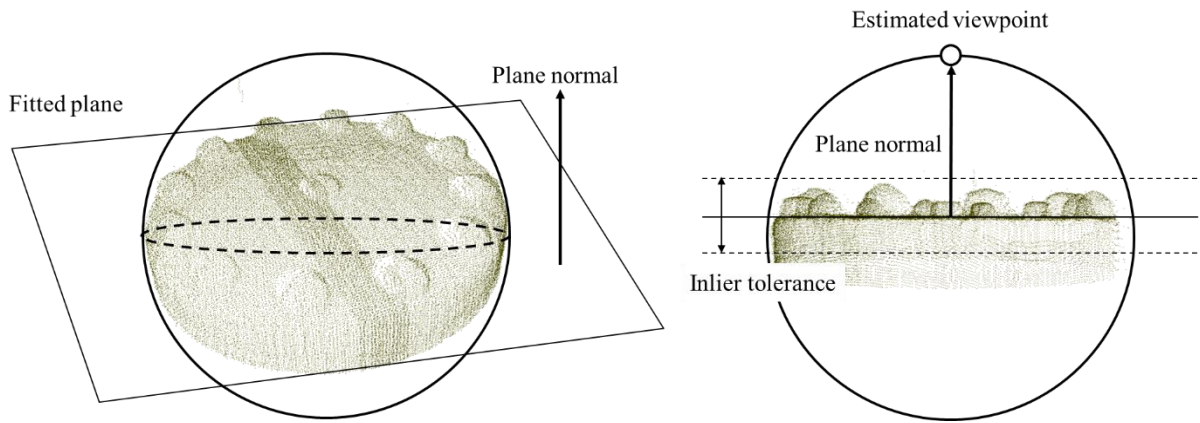

**Figure 12.** Comparative viewpoint selection method based on RANSAC

The precision-recall curve that illustrates the retrieval performance for each dataset is shown in Figures 13, 14 and 15.

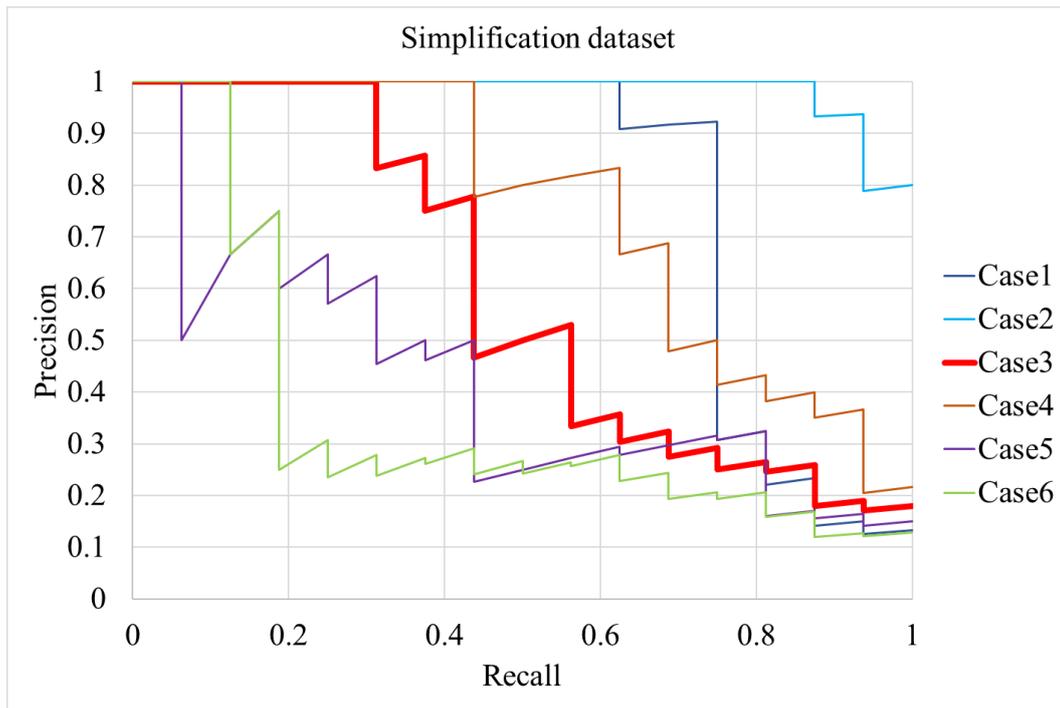

**Figure 13.** Precision-recall curves of experiments for the simplification dataset

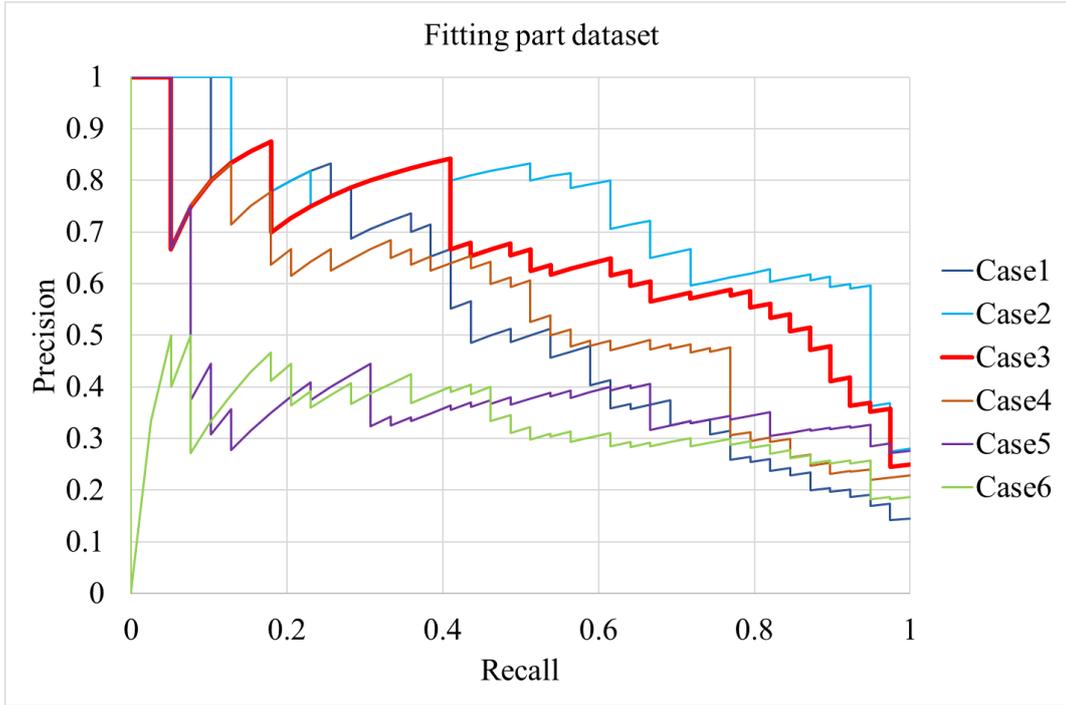

**Figure 14.** Precision-recall curve of the retrieval experiment for the fitting part dataset

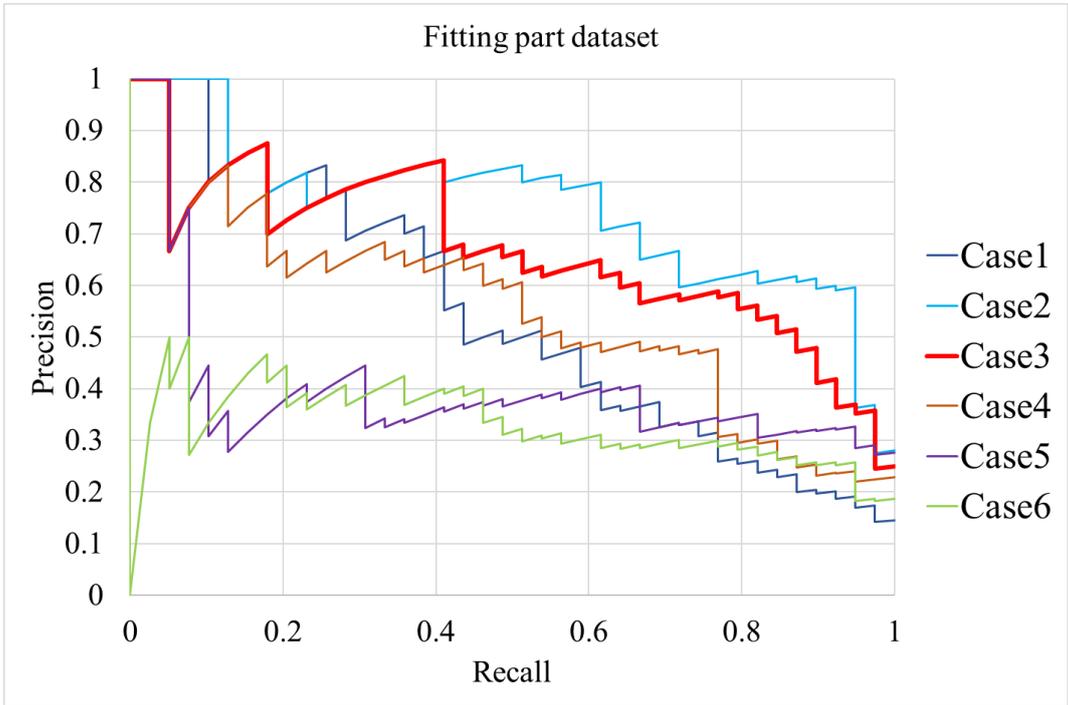

**Figure 15.** Precision-recall curve of the retrieval experiment for the segmented point cloud dataset

In the case of model simplification dataset, Case 1 demonstrated the highest performance and Cases 5 and 6 demonstrated relatively lower performances. Cases 2, 3, and 4 demonstrated similar performances. In the case of fitting part dataset, Case 3, which is the proposed method, demonstrated the highest performance after Case 2. The remaining cases demonstrated lower performance on retrieval. For both datasets, the view selection method using the RANSAC yielded the lowest performance, which indicates that the selection of viewpoint is an important factor in view-based partial shape retrieval. In the case of segmented point cloud dataset, Case 3 demonstrated the highest precision when recall was low while Case 5 showed better performance in overall.

As presented in Table 4, when comparing the angle difference between the ground truth viewpoint used to generate the point cloud and the viewpoints estimated by the RANSAC and proposed method, the viewpoint estimated by the RANSAC results in approximately 3 time greater error than the proposed method.

**Table 4.** Angular error of estimated viewpoints

|  | *Angle difference from ground truth [rad]* | |
| --- | --- | --- |
| *Viewpoint estimation* | *Simplification dataset* | *Fitting part dataset* |
| RANSAC | 0.6875 | 0.7698 |
|  | (s = 0.3385) | (s = 0.4476) |
| Proposed | 0.2238 | 0.2698 |
|  | (s = 0.1286) | (s = 0.1535) |

Table 5 shows overall performance of retrieval on dataset. The outperformed case is indicated by bold letters, except for ground truth cases (case 1 and case 2). The nearest neighbor (NN) was calculated from the results of the retrieval experiments to demonstrate the retrieval accuracy quantitatively. The nearest neighbor is the "percentage of the closest matches that belong to the same class as the query" [41]. The NN values are given as average values of all the queries for each dataset in the experiments. The results are listed in Table 5. Similar results were obtained with the analysis of precision-recall curves described above. The proposed method (Case 3) demonstrated NN of 62.50% while ground truth (Case 2) demonstrated 93.75% with the model simplification dataset. In case of fitting part dataset, NN of the proposed method and ground truth was 64.10% and 66.67%, respectively. In segmented point cloud dataset, the proposed method showed the highest performance. Mean average precision (mAP) is mean value of average precision of each class where average precision is computed by changing similarity from 0 to 1. In case of fitting part dataset, the proposed method showed 67.02% as the result. However, in simplification and segmented point cloud dataset, the proposed method without resolution estimation showed better results. Normalized discounted cumulative gain (NDCG) is a sum of true probability after applying logarithmic discount followed by normalization. Similar to other metrics, Case 3 and 4 showed better results compared to RANSAC-based estimation and known viewpoint (Case 1) in some cases.

**Table 5.** Performance metric measured for experiment results

| Dataset | Simplification | | | Fitting part | | | Segmented point cloud | | |
|---|---|---|---|---|---|---|---|---|---|
| Metric | NN | mAP | NDCG | NN | mAP | NDCG | NN | mAP | NDCG |
| Case 1 | 68.75 | 79.26 | 84.05 | 43.59 | 54.65 | 74.11 | - | - | - |
| Case 2 | 93.75 | 98.36 | 100 | 66.67 | 75.47 | 88.74 | - | - | - |
| Case 3 | 62.50 | 59.57 | 76.04 | **64.10** | **67.02** | **82.65** | **75.00** | 56.40 | **87.63** |
| Case 4 | 68.75 | **75.34** | **85.69** | 48.72 | 55.50 | 79.31 | 58.33 | **57.80** | 78.27 |
| Case 5 | 37.50 | 43.43 | 70.00 | 23.08 | 40.41 | 68.62 | 70.83 | 55.51 | 85.86 |
| Case 6 | 25.00 | 36.19 | 62.00 | 25.64 | 34.19 | 71.31 | 58.33 | 52.02 | 81.02 |

The analysis for the failure cases is described as follows. In this paper, we proposed a method of estimating the image resolution for a depth image rendered after the selection of the viewpoint. We observed that there was a tendency of underestimated density if an error occurred on selecting the viewpoint process and thus, the selected image resolution was too low. The underestimation of resolution was induced owing to a small amount of error in viewpoint, as presented in Table 4. Therefore, as evident from the experimental results for the model simplification dataset, there is a case where fixed image resolution yields better performance than the proposed resolution estimation method. However, in segmented point cloud dataset, which is a dataset collected from real world scanning, proposed method results best performance. Therefore, using the proposed method is superior in terms of automation of the retrieval process.

### 5.2 Experiment on deep-learning-based shape retrieval

Further experimentation has been conducted on the deep-learning-based shape retrieval approach. MVCNN [26] propose a network architecture where multiple images rendered from a 3D object can be recognized as class labels. To extract features from multiple images, the authors design the convolutional neural network that shares weights in first network. Next, the second layer pools important features from multiple outputs of first network by max pooling. Lastly, fully connected network classifies the object. Please refer to [26] for further details of the method.

For the experimentation of the proposed method with MVCNN shape retrieval, we extend our view selection method into multiple image rendering. Figure 16 (a) shows the multi-view rendering method proposed in MVCNN. The images are rendered around the object with elevation. The viewpoint position is sampled every 30 degrees in azimuth where elevation angle is fixed at 30 degrees. Therefore, 12 images are rendered for a single object. Figure 16 (b) illustrates the extended multi-view rendering method. The red square shows an estimated viewpoint either by the proposed method or RANSAC. To generate multi-view images, elevation and azimuth angle of viewpoint are changed by 40 degrees. Number of images for a single object is therefore 13.

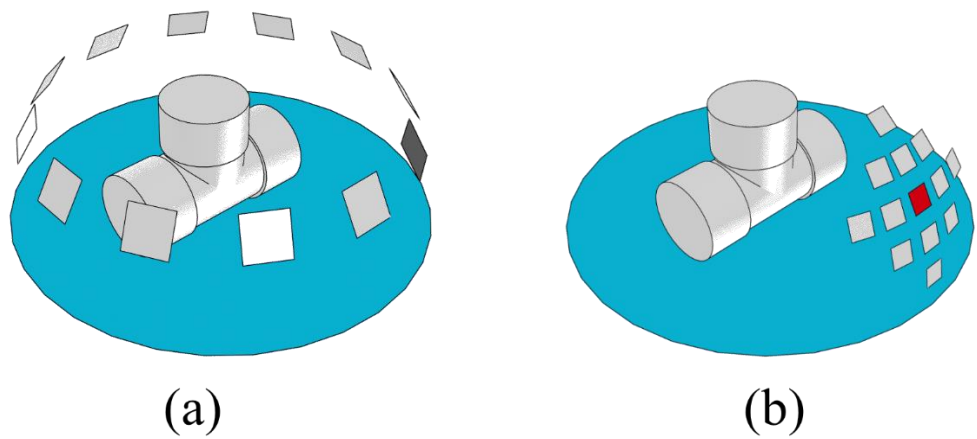

**Figure 16.** Multi-view rendering method; (a) MVCNN multi-view (b) estimated viewpoint-centered multi-view

To train the MVCNN network, we used a repository of segmented point cloud [42]. The repository consisted of 4,633 individual point clouds of 18 type fitting components. The individual point clouds were obtained from scan data of an actual process plant by manually segmenting each fitting component. The segmentation quality and type classification results were verified by domain experts.

Table 6 shows performance metrics measured for different multi-view methods. Multi-view rendering method proposed in [26] shows better performance on NN and NDCG. However, the proposed method shows better performance on mAP. A characteristic of a repository of segmented point clouds is severe class imbalance, where some classes have a lot of dataset while other classes do not. For instance, Elbow 90 class in the repository has 1,158 models but only 6 models are identified as Wye class. Since mAP is the mean value of average precision across classes, we can conclude that the proposed method-centered multi-view performs better for overall retrieval, while MVCNN multi-view is better if a sufficient amount of training data exists. Figure 17 shows precision-recall curve of the repository of segmented point cloud dataset.

**Table 6.** Performance metrics measured for MVCNN with different multi-views

| *Multi-view methods* | *NN* | *mAP* | *NDCG* |
|---|---|---|---|
| MVCNN multi-view | **87.43** | 78.55 | **93.38** |
| RANSAC-centered multi-view | 82.51 | 81.31 | 92.10 |
| Proposed method-centered multi-view | 85.03 | **81.66** | 92.35 |

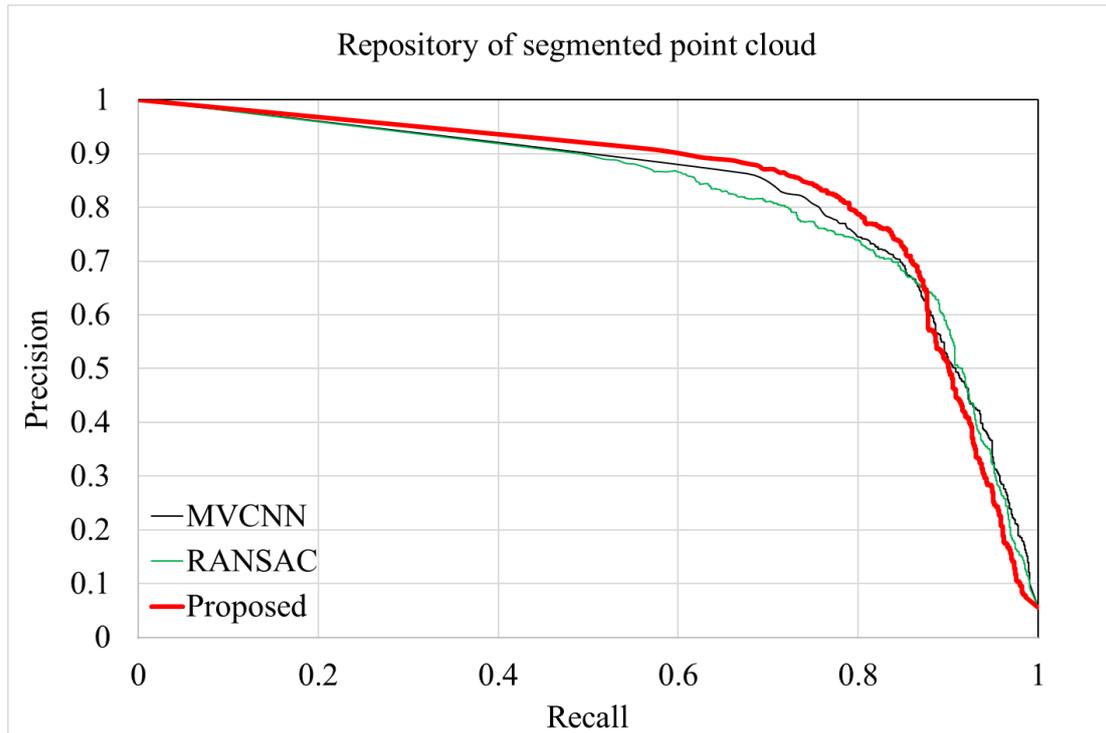

**Figure 17.** Precision-recall curve of the retrieval experiment for the segmented point cloud dataset

## 6. Conclusion

We proposed a method of selection of the viewpoint and image resolution for depth image rendering required to apply the view-based partial 3D shape retrieval to a segmented point cloud obtained from 3D scanning of large facilities. In the proposed method, multiple depth images were rendered from sampled viewpoints and sampled resolutions, and subsequently, the acquisition rate and density of information were calculated. We defined equations to select the viewpoint and image resolution. For the model simplification, fitting part and segmented point cloud datasets, the proposed method exhibited the retrieval performance of 64.10, 67.02 and 75.00 on NN and 59.57, 67.02, 56.40 on mAP, respectively. The proposed method yielded higher accuracy than the other methods it was compared to(e.g., RANSAC based viewpoint selection). In the experiment on deep-learning-based shape retrieval, the proposed method showed best performance on mAP metric. The performance showed a competitive advantage to multi-view rendering method proposed in MVCNN considering the severe class imbalance of experiment dataset. It is advantageous from the perspective of automation of the retrieval process because the proposed method does not require any input from the user when applying the view-based partial shape retrieval to a segmented point cloud of a large-scale facility. However, the retrieval accuracy of the experiment shows a difference according to the characteristics of datasets. Therefore, further research on the robustness of retrieval is required with the improvement of retrieval accuracy.

**Acknowledgment**

This research was supported by the Industrial Core Technology Development Program (Project ID: 20000725) funded by the Korea government (MOTIE), by the Basic Science Research Program (Project ID: NRF-2019R1F1A1053542 and 2020R1G1A1008932) through the National Research Foundation of Korea (NRF) funded by the Korea government (MSIT) .